\documentclass[a4paper,fleqn]{cas-dc}
\usepackage{tikz}
\usepackage{tabularx}
\usepackage{svg}
\usepackage{subcaption}
\usepackage{algorithm}
\usepackage{cleveref}
\usepackage{algpseudocode}
\usepackage{amsmath}
\usepackage[authoryear,longnamesfirst]{natbib}

\def\tsc#1{\csdef{#1}{\textsc{\lowercase{#1}}\xspace}}
\tsc{WGM}
\tsc{QE}
\tsc{EP}
\tsc{PMS}
\tsc{BEC}
\tsc{DE}

\begin{document}
\let\WriteBookmarks\relax
\def\floatpagepagefraction{1}
\def\textpagefraction{.001}


\shortauthors{A Bekkair et~al.}

\title [mode = title]{Unsupervised Graph Attention Autoencoder for Attributed Networks using K-means Loss}
\tnotemark[1,2]

\author[1,2]{Abdelfateh Bekkair}[ orcid=0009-0003-8067-0580]

\ead{bekkair.abdelfateh@univ-ghardaia.dz}

\affiliation[1]{organization={Dept. of Mathematics and Computer Science, Université de Ghardaia},
    city={Ghardaia},
    country={Algeria}}

\author[1,2]{Slimane Bellaouar} [orcid=0000-0001-8357-5501]
\ead{bellaouar.slimane@univ-ghardaia.dz}
\author[1,2]{Slimane Oulad-Naoui}[orcid=0000-0001-8357-5501 ]
\ead{ouladnaoui@univ-ghardaia.dz}


\affiliation[2]{organization={Lab. des Mathématiques et Sciences Appliquées (LMSA), Université de Ghardaia},
    city={Ghardaia},
    country={Algeria}}

\cortext[cor1]{Corresponding author}
\cortext[cor2]{Principal corresponding author}

\maketitle

\section{Proposed method}\label{prop}
\begin{figure*}
    \centering
    \input{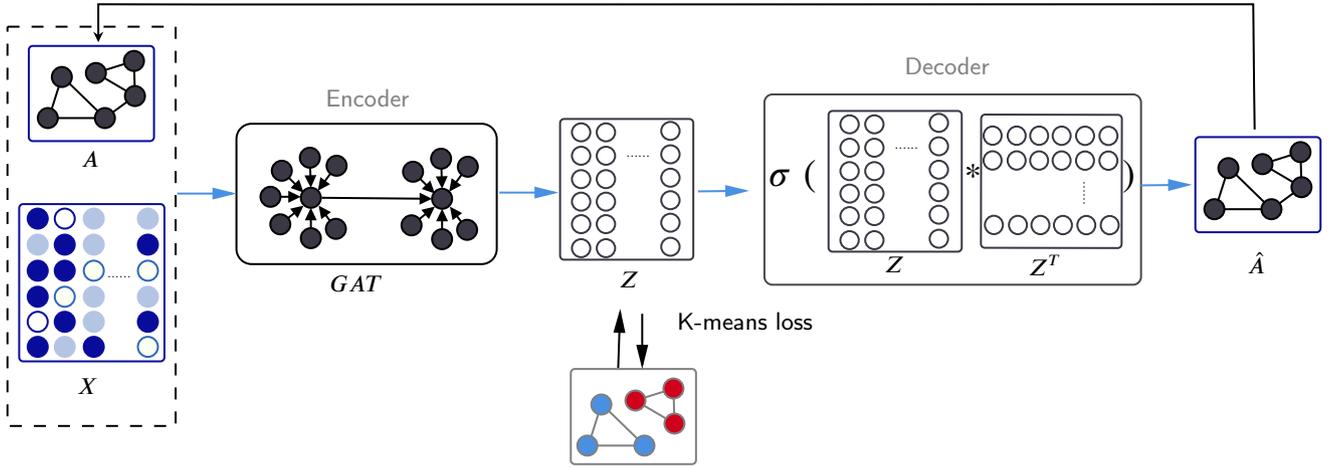}
    \caption{Architecture of the proposed GAECO model.}
    \label{fig:architecture}
\end{figure*}

In this section, we aim to illustrate the architecture of the GAECO model. The proposed model is an unsupervised Graph Attention Autoencoder, which combines a Graph Attention Network (GAT) encoder with a simple inner product decoder. The proposed architecture is visually depicted in Figure \ref{fig:architecture}. 
The autoencoder takes two matrices as input: the structural graph information, denoted as $A$, and the attribute matrix, denoted as $X$, with the aim of learning a latent representation that enables the accurate reconstruction of the topological information.
Our autoencoder encompasses three essential components: an encoder, a decoder, and a loss optimization mechanism. Here's an overview of each:
\subsection{Encoder}

The attention mechanism, introduced in the context of Graph Neural Networks (GNN), has given rise to a novel form of graph filtering within Graph Attention Networks (GATs) \citep{IntroduceAttention, GAT}.

GATs share a similarity with GCN (Graph Convolutional Networks) in the way they aggregate node features within neighborhoods to generate new features for each node. However, GAT introduces a distinguishing aspect by not solely focusing on the graph structure; it also endeavors to discern the significance of neighbors during the aggregation process. This differentiation is achieved through the utilization of trainable weights with attention mechanisms. The updated representation of node \( i \) in layer \( l+1 \) is calculated as the weighted sum of its neighbors' hidden states \( h_{j}^{(l)} \), where the attention coefficients \( \alpha_{ij}^{(l+1)} \) serve as trainable weights. This innovative approach enhances the network's ability to selectively emphasize important neighbors while incorporating their information into the aggregation:
\begin{equation}
    h_{i}^{(l+1)} = \sigma \left( \sum_{j \in N(v_i)} \alpha_{ij}^{(l+1)} W^{(l+1)} h_{j}^{(l)} \right)
\end{equation}
In this equation:
\( \sigma \) signifies the activation function, 
\( N(v_i) \) represents the neighbors of node \( i \), and 
 \( W^{(l+1)} \) signifies the trainable weight matrix for layer \( l+1 \).

To enhance the understanding of the formula, we provide the equation for calculating the attention coefficients of the neighbors of node \(i\).
\begin{equation}
\alpha_{ij}^k = \frac{\exp(e_{ij}^k)}{\sum_{m \in N(i)} \exp(e_{im}^k)}
\end{equation}
Where:
$e_{ij}^k $ is the edge-specific compatibility score between nodes $i$ and  $j$ for head  $k$. Calculated by :
\begin{equation}
\label{eq:edge}
    e_{ij}^k = \text{LeakyReLU}\left(a^k(W^k h_i, W^k h_j)\right)
\end{equation}
In Equation \ref{eq:edge}, $W^k$ is a learnable weight matrix specific to an attention head, $h_i$ and  $h_j$ are the feature vectors of nodes $i$ and $j$ respectively, and  $a^k$ is an aggregation function.

We utilize GAT as an encoder in our model for representation learning, with the aim of embedding rich and unstructured information from both the topological structure and attributed data, transforming them from their high-dimensional form within the network into a valuable low-dimensional representation denoted as \(Z\), where $Z \in \mathbb{R}^{n \times d} $. Here, $n$~and~$d$ are the number of nodes in the network and the dimension of the latent space, respectively. This transformation is achieved through the function \(\phi_{enc}(A, X)\), where \(A\) represents the structure of the graph and \(X\) corresponds to the attributed information. The outcome is a compact representation \(Z\) in a lower-dimensional space.

\subsection{Decoder}


The decoder, following the encoder phase, is responsible for generating a reconstructed representation of the input data from the encoded information. This process involves projecting the encoded information using the function $\phi_{dec}(Z)$ to obtain a reconstructed representation denoted as $\hat{A}$, which represents the topological structure. Decoders come in various types, and in our specific case, we have opted for a straightforward inner decoder, often referred to as the dot product decoder, as given in the Equation ~\ref{decoder}. This choice is driven by its reputation for efficiency and adaptability.
\begin{align}
  \hat{A} &= \sigma(Z.Z^T),
  \label{decoder}
\end{align}
Where $\sigma(\cdot)$ is a \emph{Sigmoid} activation function.
 
\subsection{Optimization}
Optimization techniques are pivotal in strengthening a deep learning model's capability to capture significant representations for community discovery tasks. Our model's objective function integrates both reconstruction optimization and joint clustering-oriented information, and these components are iteratively updated at each epoch. The optimization formulas for this purpose are as follows:
\begin{equation}
     \mathcal{L}= \mathcal{L}_{r}+\beta  \mathcal{L}_{c}.
\end{equation}

Where \(\mathcal{L}_{r}\) is  the reconstruction loss, \(\mathcal{L}_{c}\) is  the clustering loss and \(\beta\) is a positive hyperparameter for controlling the importance of the clustering optimization.

Minimizing the reconstruction error between the input data and the reconstructed data is a fundamental objective in feedforward architectures. In GAECO model, our primary goal is to minimize both the network structure topology represented by $A$ and the reconstructed adjacency denoted as $\hat{A}$, achieving this through the utilization of the binary cross-entropy loss function as follows:
\begin{equation}
    \mathcal{L}_{r} = -\frac{1}{N} \sum_{i=1}^{N} \left[ A_i \cdot \log(\hat{A}_i) + (1 - A_i) \cdot \log(1 - \hat{A}_i) \right]
\end{equation}

Additionally, our secondary objective is to shape the latent representation to achieve two primary goals: first, to minimize reconstruction error, and second, to guide the representation towards minimizing a clustering objective. To fulfill the latter, we propose the k-means loss as a method of self-learning clustering. This clustering objective helps the model discover meaningful clusters within the latent space, complementing the primary reconstruction objective.

Choosing K-means as a loss function is motivated by its simplicity and ability to create clear clusters, enhancing data representation for community discovery. It encourages grouping nodes based on similarity, aligning with the notion that nodes within a community should be more alike. Minimizing this loss aims to create compact clusters, improving the quality and interpretability of community discovery.

By incorporating K-means loss into the GAE, the model is encouraged to learn representations that not only capture the inherent structure of the data but also facilitate the natural grouping of nodes into communities. This dual objective of reconstructing input data and optimizing for community structure enhances the overall performance of the GAE, providing a unique and effective solution for community detection tasks. The K-means loss serves as a guiding force, steering the learning process toward discovering meaningful community structures within the encoded representations.


Let $Z$ be the latent representation, $C$ be the cluster centroids, and $K$ be the number of clusters.

The k-means clustering loss can be defined as the sum of squared distances between each data point in  Z and the closest centroid:
\begin{equation}
\mathcal{L}_{c} = \frac{1}{N} \sum_{i=1}^{N} \min_{j} \left\| Z_i - C_j \right\|^2
\end{equation}
Where. $\left\| \cdot  \right\|$ represents the Euclidean distance and  $C_j$ is the centroid of cluster $j$.

\section{Experiments}\label{exp}

In this section, we evaluate our approach by analyzing the results of the detected communities. We cover everything from the environment and tools used to the discussion's outcomes, including metrics, datasets, competitive models, and an ablation study. We conclude this section by presenting the results and engaging in a detailed discussion.

\subsection{Environment and Tools}
Experiments were conducted using the freely available version of Google Colab, equipped with 12.7 GB of RAM and a GPU with 15 GB of RAM. Additionally, we use the \emph{PyTorch Geometric}\footnote{https://pytorch-geometric.readthedocs.io/} library, a widely-used framework for geometric deep learning, was integrated into the Colab environment. This library provides essential tools and functionalities for working with graph-structured data, enabling seamless implementation of graph neural networks (GNN) and related techniques.

\begin{table}[width=.9\linewidth,cols=4,pos=h]
\caption{Performance on Cora.}
\label{tab:cora}
\begin{tabular*}{\tblwidth}{@{} lcll@{} }
\hline
\textbf{Model} & \textbf{Input} & \textbf{NMI}   & \textbf{ARI}   \\ \hline
K-means        & $X$              & 0.547          & 0.501          \\
DeepWalk       & $A,X$          & 0.384          & 0.291          \\
TADW           & $A,X$            & 0.366          & 0.240          \\
GAE            & $A,X$            & 0.397          & 0.293          \\
VGAE           & $A,X$           & 0.408          & 0.347          \\
MGAE           & $A,X$          & 0.511          & 0.448          \\
ARGA           & $A,X$            & 0.449          & 0.352          \\
ARVGA          & $A,X$           & 0.450          & 0.374          \\
DAEGC          & $A,X$            & 0.528          & 0.496          \\
GATE           & $A,X$           & 0.527          & 0.451          \\
ARVGA-AX       & $A,X$            & 0.526          & 0.495          \\
SENet          & $A,X$            & 0.550          & 0.490          \\
VGAER           & $A,B$           & 0.4468          & -          \\
CDBNE          & $A,X$            & 0.537          & 0.513          \\ \hline
GAECO            & $A,X$            & \textbf{0.564} & \textbf{0.516} \\ \hline
\end{tabular*}%
\end{table}

\begin{table}[width=.9\linewidth,cols=4,pos=h]
\caption{Performance on Citesser.}
\label{tab:citesser}
\begin{tabular*}{\tblwidth}{@{} lcll@{} }
\hline
\textbf{Model} & \textbf{Input} & \textbf{NMI}   & \textbf{ARI}   \\ \hline
K-means        & $X$              & 0.312          & 0.285          \\
DeepWalk       & $A,X$           & 0.131          & 0.137          \\
TADW           & $A,X$            & 0.320          & 0.286          \\
GAE            & $A,X$           & 0.174          & 0.141          \\
VGAE           & $A,X$            & 0.163          & 0.101          \\
MGAE           & $A,X$            & 0.412          & 0.414          \\
ARGA           & $A,X$            & 0.350          & 0.341          \\
ARVGA          & $A,X$            & 0.261          & 0.245          \\
DAEGC          & $A,X$            & 0.397          & 0.410          \\
GATE           & $A,X$            & 0.401          & 0.381          \\
ARVGA-AX       & $A,X$            & 0.338          & 0.301          \\
SENet          & $A,X $           & 0.417          & 0.424          \\
VGAER           & $A,B$           & 0.2169          & -          \\
CDBNE          & $A,X $           & 0.438          & 0.455          \\ \hline
GAECO            & $A,X $           & \textbf{0.451} & \textbf{0.477} \\ \hline
\end{tabular*}%
\end{table}
\begin{table}[width=.9\linewidth,cols=4,pos=h]
\caption{Performance on Pubmed.}
\label{tab:pubmed}
\begin{tabular*}{\tblwidth}{@{} lcll@{} }

\hline
\textbf{Model} & \textbf{Input} & \textbf{NMI}   & \textbf{ARI}   \\ \hline
K-means        & $X$            & 0.278          & 0.246          \\
DeepWalk       & $A,X$          & 0.238          & 0.255          \\
TADW           & $A,X$          & 0.224          & 0.177          \\
GAE            & $A,X$          & 0.249          & 0.246          \\
VGAE           & $A,X$          & 0.216          & 0.201          \\
MGAE           & $A,X$          & 0.282          & 0.248          \\
ARGA           & $A,X$          & 0.276          & 0.291          \\
ARVGA          & $A,X$          & 0.117          & 0.078          \\
DAEGC          & $A,X$          & 0.266          & 0.278          \\
GATE           & $A,X$          & 0.322          & 0.299          \\
ARVGA-AX       & $A,X$          & 0.239          & 0.226          \\
SENet          & $A,X$          & 0.306          & 0.297          \\
VGAER           & $A,B$           & 0.2129          & -          \\
CDBNE          & $A,X$          & 0.336          & \textbf{0.337} \\ \hline
GAECO            & $A,X$          & \textbf{0.341} & 0.321          \\ \hline
\end{tabular*}%
\end{table}

\subsection{Metric}
We utilize the Normalized Mutual Information (NMI) and the adjusted Rand Index (ARI) as evaluation metrics, which are widely employed for assessing community detection tasks. Improved clustering outcomes should result in elevated values across all evaluation metrics.

NMI, a pivotal metric for community detection, quantifies the agreement between the true community structure ($C$) and the detected communities ($C^*$) by measuring shared information, leveraging concepts of mutual information and partition entropy, with higher values indicating stronger agreement, as defined in Equation ~\ref{nmi}~\citep{NmiPaper}.
\begin{equation}
    \label{nmi}
NMI(C,C^*)=\frac{-2\sum_{i=1}^{k}\sum_{i=1}^{k^*} n_{ij} \log\frac{n_{ij}n}{n_i n_j} }{\sum_{i=1}^{k} n_i \log \frac{n_i}{n}+\sum_{j=1}^{k^*}n_j \log \frac{n_j}{n}}
\end{equation}
Here, $n$ denotes the count of nodes, while $k$ and $k^*$ represent the quantities of ground truth and detected communities, respectively. In this context of $n_{ij}$, it signifies the number of nodes that participate in both communities $n_i$ and $n_j$.
ARI, a commonly used statistical metric in tasks such as clustering and community detection, evaluates the agreement between the ground truth $C$ and detected communities $C^*$, offering a comprehensive view by considering both correct and incorrect groupings relative to random assignment, as expressed by Equation~\ref{ari}.

\begin{equation}
\begin{aligned}[c]
ARI(C, C^*) = \frac{{\sum_{ij} \binom{n_{ij}}{2} - \left[ \sum_i \binom{a_i}{2} \sum_j \binom{b_j}{2} \right] / \binom{n}{2}}}{{\frac{1}{2} \left[ \sum_i \binom{a_i}{2} + \sum_j \binom{b_j}{2} \right] - \left[ \sum_i \binom{a_i}{2} \sum_j \binom{b_j}{2} \right] / \binom{n}{2}}}
\end{aligned}
\label{ari}
\end{equation}

\subsection{Datasets}
We conduct experiments on three real-world datasets, specifically citation networks characterized as attributed networks. These datasets include \emph{Cora}, \emph{CiteSeer}, and \emph{PubMed}, and their brief descriptions are provided in Table~\ref{tab:datasets}~\citep{Yang2016}.

\begin{table}[width=.9\linewidth,cols=5,pos=h]
\caption{Benchmark graph datasets utilized in the experiments.}\label{tbl1}
\label{tab:datasets}
\begin{tabular*}{\tblwidth}{@{} LCCCC@{} }
\toprule
Dataset & N° Node & N° Edge & N° Features & N° Communities\\
\midrule
Cora & 2 708 &  5 429 & 1 433 & 7 \\
CiteSeer & 3 327 & 4 732 &3 703 & 6\\
PubMed & 19 717 &  44 338 & 500 & 3\\

\bottomrule
\end{tabular*}
\end{table}
\subsection{Competitive models} \label{sec:baseline} 
To evaluate the effectiveness of the proposed GAECO model, we employ a selection of well-established competitive models. We selected these models based on their prominence in the field, specifically for the task of community detection using the same network datasets. Our choices span a diverse range of methodologies, encompassing both traditional and deep models. Traditional approaches, like k-means, rely solely on the feature matrix as input. In contrast, models such as deepwalk, categorized as shallow embedding methods, are also under consideration.

Further augmenting our evaluation, we incorporate a group of deep models. These models, unlike their counterparts, take both the adjacency matrix and the feature matrix as inputs. Within this category, we have a variety of approaches like TADW, GAE, MGAE, ARGA, ARVGA, DAEGC, GATE, and CDBNE. 
Additionally, we include the VGAER model in our assessment, which takes both the adjacency matrix and the modularity matrix as inputs. All these model descriptions are succinctly summarized in Table~\ref{tab:baseline}.

\begin{table*}[!h]
\caption{Competitive models description.}
\label{tab:baseline}
\begin{tabularx}{\textwidth}{@{}p{0.4\textwidth}X@{}}

\toprule
Model                                                                                               &  Desciption\\ \midrule
K-means \citep{kmeans} & 
  Initializes K points as cluster centers and subsequently aims to minimize the intra-cluster distances among elements while simultaneously maximizing the distances between clusters. \\ \midrule
  
DeepWalk \citep{deepwalk}& 
  Employs a random walk strategy to transform a graph into a sequence of nodes, effectively capturing the underlying graph structure. This sequence is then harnessed in conjunction with Word2Vec to obtain meaningful node embeddings. \\ \midrule
  
Text-Associated DeepWalk (TADW) \citep{TADW}          &    It extends the basic DeepWalk approach by incorporating semantic information from node textual content in addition to the network's topology information                                                                                                    \\ \midrule

Graph AutoEncoder (GAE) \& Variational Graph AutoEncoder (VGAE) \citep{GAE_VGAE}    &    These models are designed as both a standard version and a variational version. Both versions utilize multi-layer graph convolutional networks to extract representations from both the adjacency and attribute matrices.                            \\ \midrule
Marginalized Graph Autoencoder for Graph Clustering (MGAE) \citep{MGAE}         &   This autoencoder utilizes Graph Convolutional Networks (GCNs). To improve representation learning, content features are intentionally corrupted with noise and efficiently handled through marginalization. By employing multiple layers of the graph autoencoder, spectral clustering is applied to the embeddings to produce the final clustering.                                                                                                  \\ \midrule
Adversarially Regularized Graph Autoencoder (ARGA) \&  Adversarially Regularized Graph Autoencoder (ARVGA) \citep{ARGA_ARVGA}&   Incorporating adversarial training into both the standard graph autoencoder and its variational counterpart is utilized to elevate the effectiveness of learning representations for graph structures.                                                                                                \\ \midrule
Deep Attentional Embedded Graph Clustering (DAEGC) \citep{DAEGC}         &    They leverage the Graph Attention Network (GAN) framework to acquire object vector representations and perform network clustering simultaneously. This is achieved through integrating reconstructed loss and killerback loss as self-clustering objectives.                                                                                                  \\ \midrule
Graph ATtention autoEncoder (GATE) \citep{GATE}       &  It is a graph auto-encoder that incorporates stacked encoder and decoder layers. The model achieves this by utilizing self-attention mechanisms to reconstruct both node features and the graph structure.                                                                                   \\ \midrule
Spectral Embedding Network for attributed graph clustering (SENet) \citep{SENet}       &  It enhances the graph structure by leveraging shared neighbor insights and refines node embeddings through a spectral clustering loss. By fusing the original graph structure with the similarity derived from shared neighbors, it integrates both structural and feature-based data into the kernel matrix using higher-order graph convolutions.                                                                                                     \\ \midrule
Variational Graph AutoEncoder Reconstruction based community detection (VGAER) \citep{VGAER}              &  By employing a variational graph autoencoder constructed using the graph convolutional network, the incorporation of network structure and modularity information into the learning process is achieved. \\ \midrule
Community Detection algorithm Based on unsupervised  attributed Network Embedding (CDBNE) \citep{CDBNE}     & Utilizes a Graph Attention Autoencoder (GAE) to derive embeddings for attributed networks through unsupervised learning. This process integrates a module for modularity maximization and a self-training clustering module, collectively enhancing the network representation. \\ \bottomrule
\end{tabularx}%
\end{table*}

\subsection{Ablation analysis}
To assess the model's effectiveness, we first provide consistent results to investigate the impact of ablation. This involves conducting experiments with and without the clustering loss component, examining how it influences the outcomes. Additionally, we explore the effect of the hyperparameter $\beta$ on our model's performance.

\begin{figure*}[ht]
	\centering
	\includegraphics[width=\textwidth]{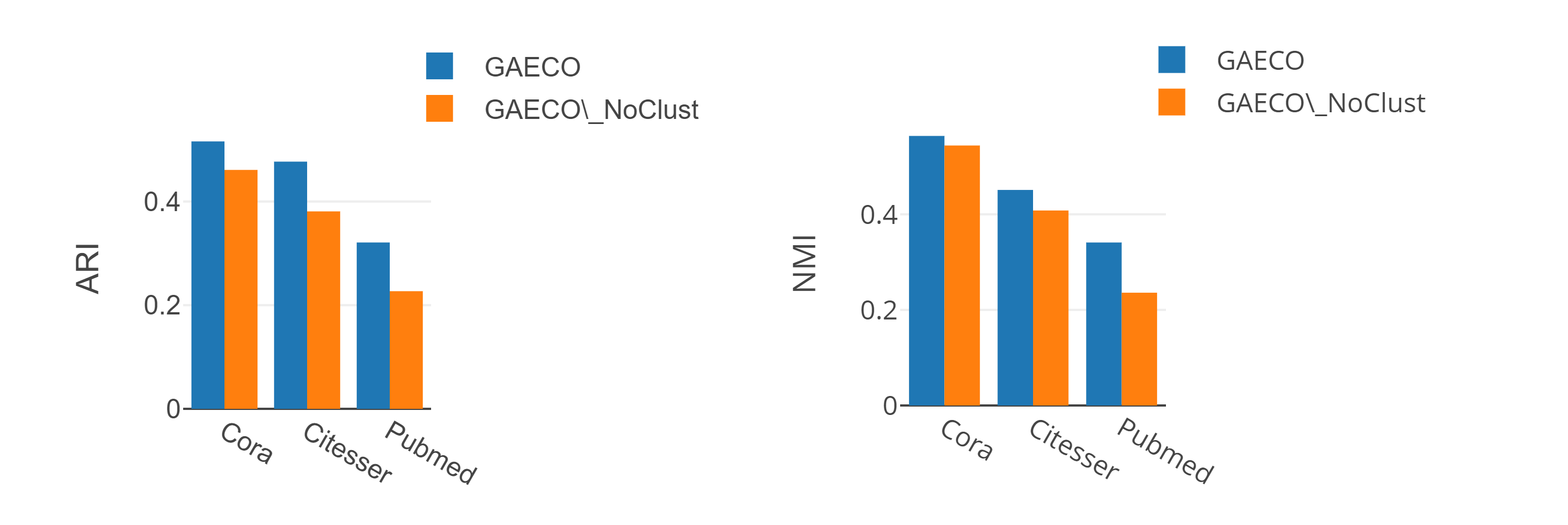}
	\caption{Ablation analysis results.}
	\label{fig:ResultAblation}
\end{figure*}

\begin{figure}[ht]
  \includegraphics[width=0.9\textwidth]{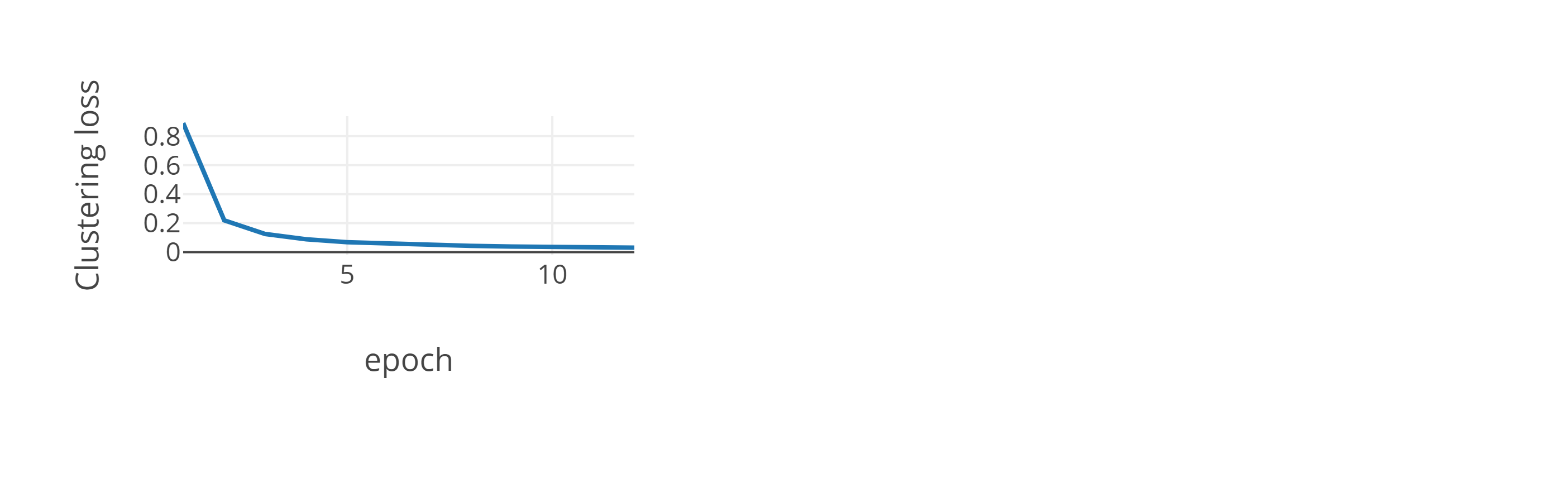}
  \caption{Clustering loss value in pubmed.}
  \label{fig:LossValue}
\end{figure}

\begin{figure}[ht]
  \centering
  \includegraphics[width=\textwidth]{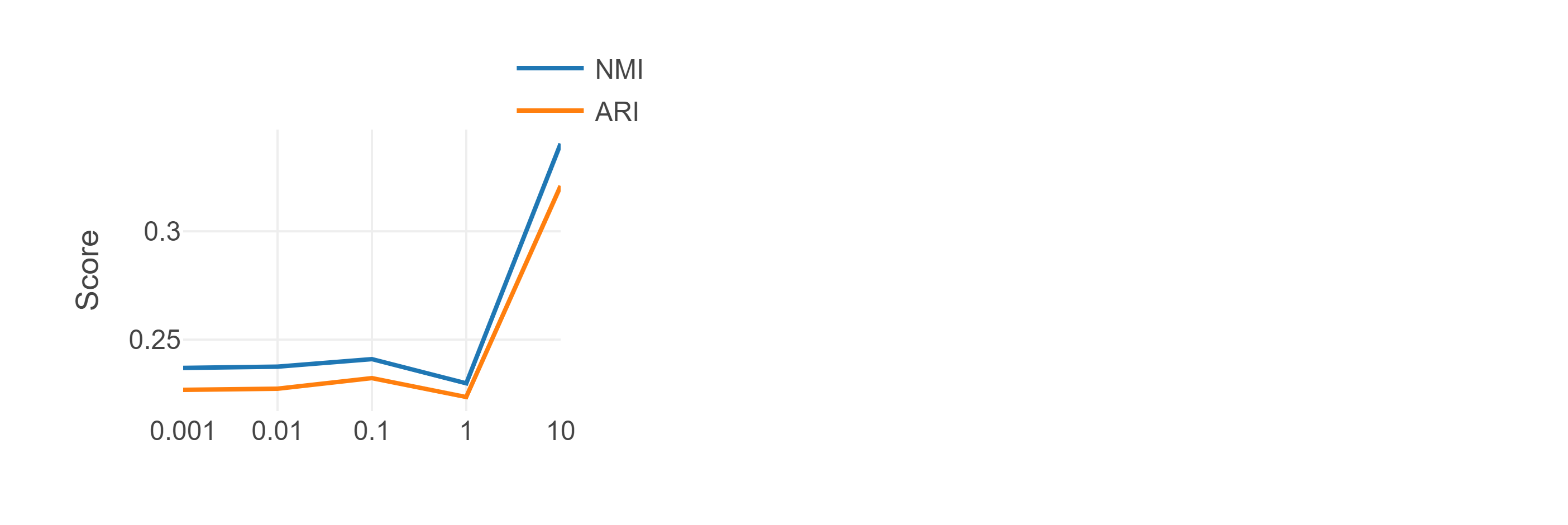}
  \caption{Different $\beta$ value result (Pubmed dataset).}
  \label{fig:BetaValue}
\end{figure}

To evaluate the influence of clustering effectiveness, Figure \ref{fig:ResultAblation} presents the NMI and ARI outcomes for \emph{GAECO} with and without the clustering loss across all the network datasets employed in this experiment. The model without clustering loss is denoted as GAECO\_NoClust. We demonstrate that the clustering loss improves the model by up to 10\% when compared to GAECO\_NoClust in terms of evaluation metrics. Furthermore, in Figure~\ref{fig:LossValue}, we illustrate how the clustering loss value decreases very gradually over training epochs. For a better understanding of the $\beta$ hyperparameter's role, Figure~\ref{fig:BetaValue} depicts how it evolves during training and its effect on NMI and ARI scores on the Pubmed dataset. It is worth noting that the best performance of the model on this dataset is achieved when $\beta$ reaches a value of 10. This parameter determines how much importance is given to the clustering loss information to improve the results in community detection.


In our implementation, we utilize an encoder architecture that consists of a hidden layer containing 256 neurons and an embedding size of 64 neurons. This architecture is employed across all three datasets, and we incorporate an 8-headed attention mechanism. For the optimization process, we employ the Adam optimizer. Regarding the clustering $\beta$ hyperparameter in the cora, citesser, and pubmed datasets, we specifically choose values of 10, 0.1, and 0.1, respectively. 
In the two layers of the GAT encoder, we employ the dropout mechanism with varying dropout rates as a means of enhancing the robustness and regularization of the model.
\subsection{Results and discussion}
The performance of the detected communities on the three datasets, Cora, Citesser, and Pubmed, is evaluated using two robust measures: NMI and ARI scores. Tables~\ref{tab:cora},~\ref{tab:citesser}~and~\ref{tab:pubmed} display the results. In addition to the metric values presented in the tables, these tables also provide information about the input data, denoted as $X$ for the attribute information, $A$ for the topology structure information, and $B$ for modularity information. The best-performing values in the tables are highlighted in bold, demonstrating that GAECO clearly outperforms all the competitive models across the evaluation metrics.

In terms of NMI scores, GAECO outperforms the best competitive model (CDBNE) by a margin of at least 3\% for Cora, 2\% for Citesser, and 1\% for Pubmed. Regarding ARI scores, GAECO model surpasses CDBNE by at least 0.1\% for Cora and 2\% for Citesser, except for Pubmed, where our model's score is slightly lower, differing by -0.1\%.

In the three tables ~\ref{tab:cora},~\ref{tab:citesser}~and~\ref{tab:pubmed} of results, we present models based on Graph Attention Autoencoder (DAEGC), GATE, CDBNE, and GAECO. All these models utilize the same input data, both A and X. These models consistently achieve superior classification results when compared to other competitive models. This suggests that the utilization of the GAT encoder is an effective choice for extracting nonlinear information.

Our model is designed with a simple yet effective architecture that follows a straightforward approach without the complexity of using multiple reconstruction layers. It consists of a Graph Autoencoder (GAE) with a multi-head Graph Attention Network (GAT) encoder and an inner product decoder. Our model is trained using a dual objective approach, which involves both reconstruction and clustering losses, with k-means employed as the clustering loss method. At the end of the training process, the k-means algorithm is applied to obtain the detected labels. In the next subsection, we will provide visualizations of both the representations and the identified communities.
\subsubsection{Visualization}

\begin{figure*}[ht]
    \centering
\begin{subfigure}[b]{0.3\textwidth}
         \centering
         \includegraphics[width=\textwidth]{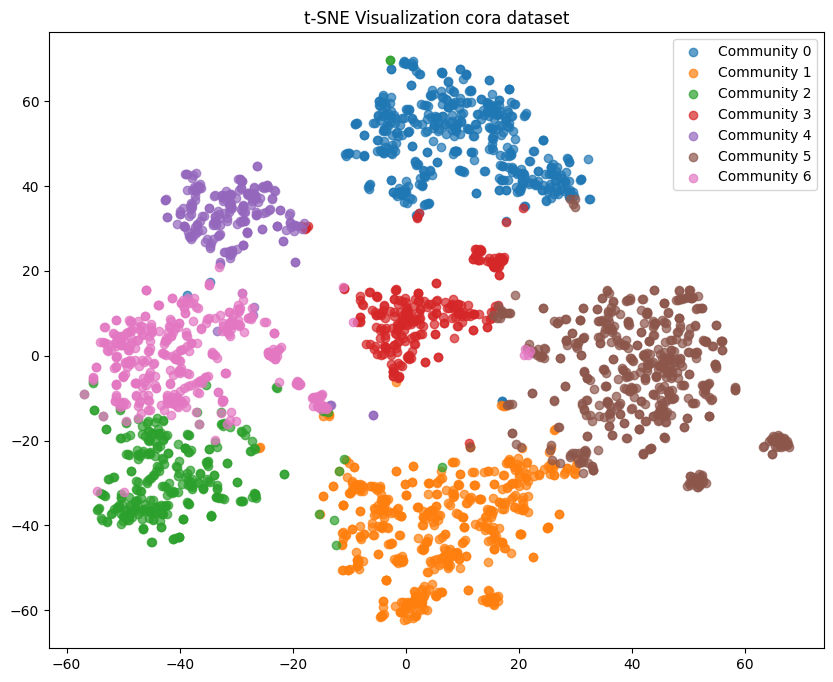}
         \caption{Cora.}
     \end{subfigure}
     \hfill
     \begin{subfigure}[b]{0.3\textwidth}
         \centering
         \includegraphics[width=\textwidth]{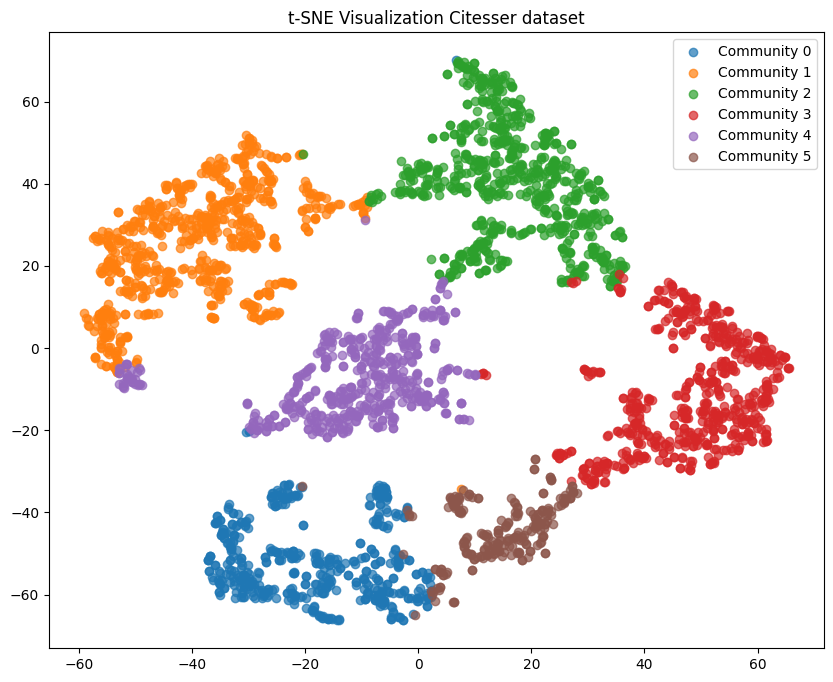}
         \caption{Citesser.}
     \end{subfigure}
     \hfill
     \begin{subfigure}[b]{0.3\textwidth}
         \centering
         \includegraphics[width=\textwidth]{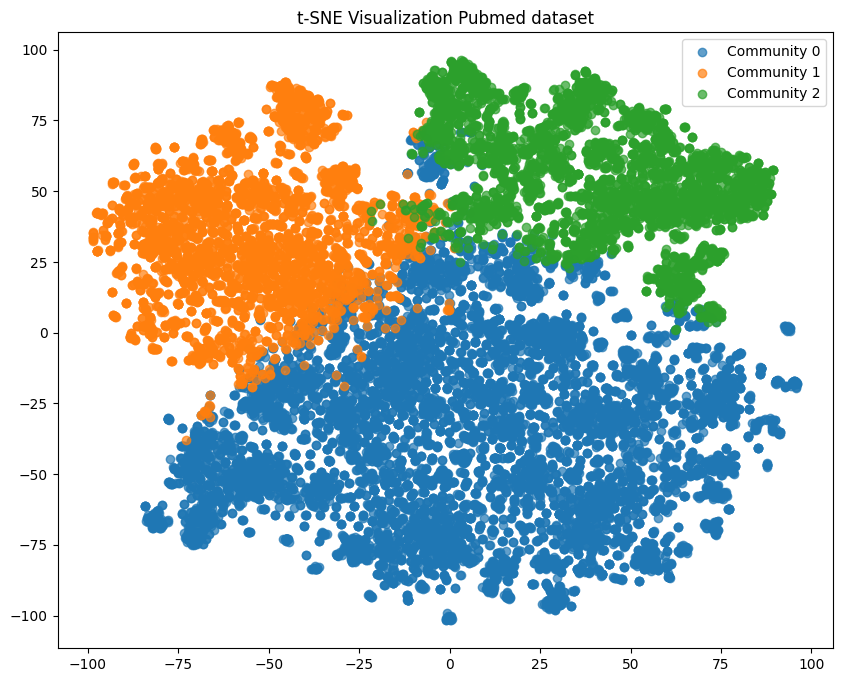}
         \caption{Pubmed.}
     \end{subfigure}
        \caption{2D visualization using t-SNE.}
        \label{fig:visualization}
\end{figure*}

In this part, our objective is to visually present and provide an intuitive understanding of the communities detected by GAECO model, which is based on the GAT (Graph Attention Network) encoder and clustering-oriented information. This understanding is achieved by learning representations derived from both the graph structure and attribute information. We project these representations into a 2D space, enabling us to display the intuitive outcomes of the detected communities across all the datasets used in our study. It's important to note that the different colors of the points in the projection space determine the node community labels. These visualizations highlight the relationships among nodes in the network, representing each community as a distinctive pattern. They also show the extent to which these communities overlap with each other, providing a clear and visually intuitive representation of the detected communities. The visualizations are presented in Figure \ref{fig:visualization}, and they are obtained using the \emph{T-SNE} algorithm \citep{TSNE} with two components.

In this paper, we introduce an unsupervised graph attention autoencoder designed for learning representations from attributed networks, with the specific goal of enhancing community detection. We propose the use of the K-means algorithm as a loss function in conjunction with the reconstruction loss to aid in generating embedding representations from both the network's topology structure and attribute information. This approach is oriented towards the clustering objective while preserving maximal structural, attribute, and relational information in the low-dimensional space.
In experiments conducted on three well-known citation networks (Cora, Citeseer, and Pubmed), our results illustrate that GAECO outperform the competitive models.
In our ongoing research, we plan to extend GAECO to address the challenges posed by large-scale networks and dynamic networks that undergo continuous changes.
\section{Acknowledgments}
\begin{sloppypar}This work is supported by the Directorate-General for Scientific Research and Technological Development (DGRSDT)-Algeria, and performed under the PRFU Project: C00L07UN470120230001. \end{sloppypar}





\printcredits

\bibliographystyle{cas-model2-names}

\bibliography{cas-refs}

\end{document}